\documentclass[aps,preprint,groupedaddress]{revtex4}
\usepackage{amsfonts}
\usepackage{mathrsfs}
\usepackage{graphicx}
\usepackage{subfigure}
\usepackage{setspace}
\usepackage{amsmath}
\usepackage{amssymb}
\usepackage{amsbsy}
\usepackage{bm}
\usepackage{algorithm}
\usepackage{physics}
\usepackage{algpseudocode} 
\newcommand{\half}{\frac12}
\newcommand{\E}{\mathscr{E}}
\begin{document}
\title{Meta predictive learning model of languages in neural circuits}
\author{Chan Li$^{1,3}$}
\thanks{Equal contribution.}
\author{Junbin Qiu$^{1}$}
\thanks{Equal contribution.}
\author{Haiping Huang$^{1,2}$}
\email{huanghp7@mail.sysu.edu.cn}
\affiliation{$^{1}$PMI Lab, School of Physics,
Sun Yat-sen University, Guangzhou 510275, People's Republic of China}
\affiliation{$^{2}$Guangdong Provincial Key Laboratory of Magnetoelectric Physics and Devices, Sun Yat-sen University, Guangzhou 510275, People’s Republic of China}
\affiliation{$^{3}$Department of Physics, University of California, San Diego, 9500 Gilman Drive, La Jolla, CA 92093, USA}
\date{\today}
\begin{abstract}
Large language models based on self-attention mechanisms have achieved astonishing performances not only in natural language itself, but also in a variety of tasks of different nature. However, regarding processing language, our human brain may not operate using the same principle. Then, a debate is established on the connection between brain computation and artificial self-supervision adopted in large language models. One of most influential hypothesis in brain computation is the predictive coding framework, which proposes to minimize the prediction error by local learning. However, the role of predictive coding and the associated credit assignment in language processing remains unknown. Here, we propose a mean-field learning model within the predictive coding framework, assuming that the synaptic weight of each connection follows a spike and slab distribution, and only the distribution, rather than specific weights, is trained. This meta predictive learning is successfully validated on classifying handwritten digits where pixels are input to the network in sequence, and moreover on the toy and real language corpus. Our model reveals that most of the connections become deterministic after learning, while the output connections have a higher level of variability. The performance of the resulting network ensemble changes continuously with data load, further improving with more training data, in analogy with the emergent behavior of large language models. Therefore, our model provides a starting point to investigate the connection among brain computation, next-token prediction and general intelligence.
\end{abstract}
 \maketitle

\section{Introduction}
Large language models (LLMs) based on transformer structures greatly boost both industrial and academic interests in artificial general intelligence~\cite{Sparks-2023}. LLMs are able to achieve state-of-art performances in a variety of different tasks, only trained by next-token prediction. The transformer structure computes self-attention scores to capture statistical correlations among input tokens in parallel, which is in stark contrast to brain-like recurrent computation based on synaptic feedback in temporal depth (e.g., a short working memory). In addition, LLMs typically require a sizable number of corpus to trigger emergence of intelligence, compared to the the fact that much less data is needed for a child to acquire linguistic ability. Therefore, it is necessary to establish a mechanistic model of language processing to understand the biological plausible mechanism and underlying physics law governing phase transitions, through statistical patterns of model hyperparameters~\cite{Huang-2023}. 

In brain science, predictive coding is one of the most influential hypothesis that can implement hierarchical information processing~\cite{PC-1999,Rao-2011}. The predictive coding derives the neuroplasticity rule based on local error signal~\cite{NC-2017},  whose goal is to minimize the surprise between the prediction and belief of a generative model of the outside world~\cite{PCrev-2021}. The framework of predictive coding has several benefits for theoretical research. First, the framework can be derived from the first principle that the brain is a biological machine of optimizing neural dynamics and synaptic connections to maximize the evidence of its internal model of the outside world~\cite{Friston-2018}. Second, this principle shares exactly the same spirit adopted in variational free energy framework~\cite{PCrev-2021}. Recently, there appeared
intense interests in studying the biological implementation of this hypothesis~\cite{Neuron-2012, Sej-2022, Wang-2023}, in developing algorithmic applications~\cite{PC-2022nips,Exact-2021,Beren-2022}, and in studying the trade-off between energy minimization and information robustness in a linear model of lateral predictive coding~\cite{Zhou-2023}. 

The predictive coding postulates that the cortex carries out a predictive model, where the incoming sensory signals are predicted by using prediction-error driven learning and inference. In this sense, predictive coding is a nice framework to model language processing. However, weight uncertainty is commonly observed in neural circuits~\cite{Pouget-2013,Taro-2021}, e.g., synaptic transmission is stochastic, and spine size is subject to fluctuation.  But these effects were not taken into account in previous studies of predictive coding, as remarked in a recent review~\cite{Brain-2023}. In addition, the weight uncertainty was recently studied in recurrent neural networks~\cite{Zou-2023}, inspiring fluctuation-driven synaptic plasticity. Therefore, exploring how the weight uncertainty affects predictive coding in language processing, will help to establish a mechanistic model of language processing to understand the biological plausible mechanism and underlying physics law governing phase transitions, through associated statistical patterns of model hyperparameters. In this work, we derive a mean-field learning rule for predictive coding in recurrent neural networks (RNNs), which is a fundamental structure for nature
language processing~\cite{LSTM-1997,Bengio-2003,Seq-2014,Cho-2014,Chung-2014,Bah-2015,Cell-2019},  and we assume that each direction of connection follows a weight distribution incorporating weight sparsity and variance. We thus call this rule meta predictive learning (MPL). This framework is tested first on the classification of MNIST dataset~\cite{mnist} where pixels in an image are divided into groups, and then a toy language dataset, where we can have a thorough exploration of algorithmic capabilities, and finally a language corpus in the real world (Penn Treebank corpus~\cite{data-1993}).

Our proposed MPL achieves equal or even better performance compared with traditional methods in all three tasks, showing the advantage of \textit{ensemble} predictive coding, since examples of single networks can be readily sampled from the trained distribution~\cite{Li-2020,Zou-2023}. By analyzing the distribution of hyperparameters, we are able to find that most connections are deterministic in the input and recurrent layer, while the output layer has a higher level of variability. The observation that the output connections bear a higher level of variability is a universal result in all three tasks,  which may particularly connect to the generative function of the language processing model. The network performance changes non-linearly and continuously with data load $\alpha = \frac{M}{N}$, where $M$ is the training data size and $N$ is the number of neurons in the circuit, and we found that the critical point is given by $\alpha_c\approx 0.02$, beyond which a chance level of prediction is absent. With increasing the size of training data, the performance further improves until a perfect learning is achieved. We can then test the resulting network to generate text of arbitrary length (to create something is a first step to understand that thing), and the generated text follows perfectly the grammatical rule set before training. In addition, our MPL is able to accomplish comparable performances in the Penn Treebank corpus with other training methods in RNN, although the framework is less accurate than the transformer structure, which thereby calls for further studies about the mechanistic difference between biological learning and non-biological transformer learning, and how the latter one can inspire discovery of new fundamental elements of computation that can realize logical and mathematical reasoning in many different tasks~\cite{NLP-2022,Tenen-2023}.

\section{Method}
Here we consider meta predictive learning in a vanilla recurrent neural network (RNN), which processes a time-dependent sequence $\mathbf{x}$ with time length $T$. The input signal of $N_{\rm in}$ dimension is first mapped to the recurrent reservoir of $N$ neurons by an input weight matrix $\mathbf{w}^{\rm in}\in \mathbb{R}^{N\times N_{\rm in}}$, whose element $w^{\rm in}_{ij}$ indicates the connection weight value from neuron $j$ in the input to the reservoir neuron $i$.  The neurons in the reservoir interact with each other with reciprocal connections $\mathbf{w}\in\mathbb{R}^{N\times N}$, where elements $w_{ij}$ specify the directional coupling from neuron $j$ to neuron $i$, and moreover $w_{ij}\neq w_{ji}$. The self-connectivity $w_{ii}$ is also included, and can be learned without imposing any prior knowledge~\cite{Zou-2023}. The internal neural dynamics $r_i(t)$ is read out via the output weight $\mathbf{w}^{\rm out}\in \mathbb{R}^{N_{\rm out}\times N}$. In the predictive learning setting, $\mathbf{r}$ is interpreted as a \textit{belief} state when $\mathbf{x}$ is observed as a sensory input, which can be continuously updated to match the actual prediction whose dynamics reads as follows, 
\begin{equation}\label{dynamics}
	\begin{aligned}
		& h_{i}(t)=\sum_{j=1}^{N} w_{i j} f\left(r_{j}(t-1)\right)+\sum_{j=1}^{N_{\rm in}} w_{i j}^{\rm in} x_{j}(t),\\
		& y_i(t)=\phi\left(\sum_{j=1}^N w_{ij}^{\rm out} f\left(r_{j}(t)\right)\right),
	\end{aligned}
\end{equation}
where $y_i(t)$ is the $i$-th component of the network output, $f(\cdot)$ denotes the non-linear activation function, and we use the ReLU function for all tasks. $\phi(\cdot)$ is the output nonlinear function, and we use softmax function to specify the probability over all classes, which is defined as $\phi(z_k(t))=\frac{e^{z_k(t)}}{\sum_j e^{z_j(t)}}$. The belief state $\mathbf{r}(t)$ is updated for all time steps up to the sequence length to minimize the prediction error between $\mathbf{r}$ and $\mathbf{h}$, which will be detailed below.
 Note that $\mathbf{r}(0) = 0$, and we fix the belief of the output node $\mathbf{r}_y = \hat{\mathbf{y}}$, where $\hat{\mathbf{y}}$ denotes the label of input $\mathbf{x}$. Generally speaking, all beliefs can be initialized to random values.

The core idea of the proposed MPL is assuming the distribution of the network parameters is subject to the following spike and slab (SaS) form~\cite{Li-2020,Zou-2023},
\begin{equation}
	\small
\begin{aligned}
	P\left(w_{i j}^{\text {in }}\right) & =\pi_{i j}^{\mathrm{in}} \delta\left(w_{i j}^{\mathrm{in}}\right)+\left(1-\pi_{i j}^{\mathrm{in}}\right) \mathcal{N}\left(\frac{m_{i j}^{\text {in }}}{N_{\rm in}\left(1-\pi_{i j}^{\rm in}\right)}, \frac{\Xi_{i j}^{\mathrm{in}}}{N_{\rm in}\left(1-\pi_{i j}^{\rm in}\right)}\right), \\
	P\left(w_{i j}\right) & =\pi_{i j} \delta\left(w_{i j}\right)+\left(1-\pi_{i j}\right) \mathcal{N}\left( \frac{m_{i j}}{N\left(1-\pi_{i j}\right)}, \frac{\Xi_{i j}}{N\left(1-\pi_{i j}\right)}\right), \\
	P\left(w_{k i}^{\text {out }}\right) & =\pi_{k i}^{\text {out }} \delta\left(w_{k i}^{\text {out }}\right)+\left(1-\pi_{k i}^{\text {out }}\right) \mathcal{N}\left( \frac{m_{k i}^{\text {out }}}{N\left(1-\pi_{k i}^{\text {out }}\right)}, \frac{\Xi_{k i}^{\text {out }}}{N\left(1-\pi_{k i}^{\text {out }}\right)}\right).
\end{aligned}
\end{equation}
Note that $N(1-\pi)$ specifies the mean degree of each neuron in the reservoir, as $1-\pi$ specifies the synaptic connection probability, which is biological plausible due to unreliable stochastic noise~\cite{Noise-2008}.
The first and second moments of elements $w_{ij}^\ell$ ($\ell$ is $\text{in}, \text{out}$, or $\text{recurrent}$ depending on the context; for the recurrent context, the item has no super- or subscript) can be derived as $ \mu^{\ell}_{i j}=\frac{m^{\ell}_{i j}}{N_\ell},  \varrho^{\ell}_{i j}=\frac{\left(m^{\ell}_{i j}\right)^2}{N_\ell^2\left(1-\pi^{\ell}_{i j}\right)}+\frac{\Xi^{\ell}_{i j}}{N_\ell}$, respectively. Note that the mean and variance of the Gaussian slab are scaled by the number of mean synaptic connections, such that the prediction of neuron is of the order one. 

Considering the statistics of synaptic weights and a large number of afferent projections for each neuron in Eq.~\eqref{dynamics}, which is true in real neural circuits~\cite{Luo-2021}, we can reasonably assume the prediction $h_i(t)$ $(\forall i)$ follows an evolving Gaussian distribution whose mean and variance are defined by $G_i=G_i^{\rm rec}+G_i^{\rm in}$ and $\Delta_i^1=(\Delta_i^{\rm in})^2+(\Delta_i^{\rm rec})^2$, respectively. This is intuitively the result of central limit theorem. The statistics of readout neural currents can be derived in a similar way. Therefore,
the mean-field dynamics of this model can be written as
\begin{equation}\label{MFD}
	\small
	\begin{aligned}
		&h_{i}(t+1) =G_{i}^{\mathrm{rec}}(t)+G_i^{\mathrm{in}}(t+1)+\epsilon_i^{\mathrm{1}}(t+1) \sqrt{\left(\Delta_i^{\mathrm{in}}(t+1)\right)^2+\left(\Delta_{i}^{\mathrm{rec}}(t)\right)^2}, \\
		&y_{k}(t) =\phi\left(G_{k}^{\text {out }}(t)+\epsilon_k^2(t) \Delta_{k}^{\text {out }}(t)\right),
	\end{aligned}
\end{equation}
where the superscript in $\epsilon$ indicates different types of standard Gaussian random variables---one for reservoir neurons (with superscript 1) and the other for readout neurons (with superscript 2). By definition, $\{\epsilon^{1,2}_i(t)\}$ are both time and neuron index dependent. Given $\mu_{ij}^\ell$ and $\varrho^{\ell}_{ij}$, the mean currents together with the associated fluctuations are derived below,
\begin{equation}
	\begin{aligned}
	G_i^{\mathrm{in}}(t+1) & =\sum_j \mu_{i j}^{\mathrm{in}} x_j(t+1) \\
	G_{i}^{\mathrm{rec}}(t+1) & =\sum_j \mu_{i j} f\left(r_{j}(t+1)\right) \\
	G_{k}^{\mathrm{out}}(t+1) & =\sum_j \mu_{k j}^{\text {out }} f\left(r_{j}(t+1)\right) \\
	\left(\Delta_i^{\mathrm{in}}(t+1)\right)^2 & =\sum_j\left(\varrho_{i j}^{\text {in }}-\left(\mu_{i j}^{\mathrm{in}}\right)^2\right)\left(x_j(t+1)\right)^2 \\
	\left(\Delta_{i}^{\mathrm{rec}}(t+1)\right)^2 & =\sum_j\left(\varrho_{i j}-\left(\mu_{i j}\right)^2\right)\left(f\left(r_{j}(t+1)\right)\right)^2 \\
	\left(\Delta_{k}^{\text {out }}(t+1)\right)^2 & =\sum_j\left(\varrho_{k j}^{\text {out }}-\left(\mu_{k j}^{\text {out }}\right)^2\right)\left(f\left(r_{j}(t+1)\right)\right)^2 .
	\end{aligned}
\end{equation}

The prediction dynamics (Eq.~\eqref{dynamics} or Eq.~\eqref{MFD} in the meta learning context) can be interpreted as perceptual inference, widely used in energy-based optimization of brain dynamics~\cite{Friston-2018}, while the learning given below is called the neuroplasticity. Both processes minimize exactly the same energy (or variational free energy in general~\cite{NC-2017}). 

Predictive learning can be derived from a temporally hierarchical Gaussian probabilistic principle~\cite{NC-2017,Rao-2011}, where the objective function is given by the negative log-likelihood of the joint neural-state distribution. To optimize this objective function, we apply a mean-field approximation of the joint distribution and an additionally Laplace approximation that leads to Gaussian forms~\cite{Brain-2023}. We give a brief interpretation in appendix~\ref{app-pc}. In essence, the predictive learning maximizing this log-likelihood aims to minimize the following energy cost~\cite{Exact-2021,Rose-2022}, 
\begin{equation}
	\mathcal{F} = \sum_{j=1}^2\sum_{t=1}^T{\E}_{j}(t).
\end{equation}
This energy function is exactly the variational free energy in the above Gaussian probabilistic principle.
The choice of $\E_j(t)$ depends on the problem at hand. If the network produces an output at every time step as in the language model, ${\E}_1(t) =\frac{1}{2}\|\mathbf{{r}}(t) - \mathbf{{h}}(t)\|^2$, and ${\E}_{2}(t) = - \sum_i{\hat{y}_i}(t)\ln\left({{y}_i}(t)\right)$. However, if the network only makes the final decision in the last time step as in the classification task, i.e., $y_{k}(T) =\phi\left(G_{k}^{\text {out }}(T)+\epsilon^2_k(T) \Delta_{k}^{\text {out }}(T)\right)$, we then have the energy terms ${\E}_1(t)= \frac{1}{2} \|\mathbf{{r}}(t) - \mathbf{{h}}(t)\|^2$ for $t=1, \ldots , T-1$, ${\E}_1(T) = 0$, and $\E_2(t)=0$ for $t<T$, ${\E}_2(T) =  - \sum_i{\hat{y}_i}\ln\left({{y}_i}(T)\right)$.
Moreover, we define the prediction error $\boldsymbol{\E}^{\prime}_1(t) = \mathbf{{r}}(t) - \mathbf{{h}}(t)$ and $\boldsymbol{\E}^{\prime}_2(t) = \mathbf{\mathbf{r}_{y}}(t) - \mathbf{{y}}(t)$. This error can be propagated along the dendritic connections in neural circuits~\cite{Senn-2018}. In mathematical sense, the prediction errors can be interpreted as the gradients of the above energy cost.

In essence, the predictive learning consists of three phases: inference phase, learning phase, and prediction phase (see Eq.~\eqref{dynamics}, and in the current meta-learning, Eq.~\eqref{MFD} is used). We next show the predictive learning details for the language processing, while other applications can be readily adapted. First of all,  during the inference phase, the belief $\mathbf{r}(t)$ is updated to minimize the energy function $\mathcal{F}$ with the following increment,
\begin{equation}\label{inference}
	\small
\begin{aligned}
\Delta r_{i}(t^{\prime}) & =-\gamma \frac{\partial \mathcal{F}}{\partial r_{i}(t^{\prime})} \\
& =-\gamma \frac{\partial \E_{1}(t')}{\partial r_i (t^{\prime})}-\gamma \frac{\partial \sum_{t \neq t^\prime}  \E_{1}(t)}{\partial r_i (t^{\prime})} -\gamma \frac{\partial \E_{2}(t')}{\partial r_i (t^{\prime})}\\
& =-\gamma {\E}^{\prime}_{1,i}(t^{\prime})+\gamma \sum_j {\E}^{\prime}_{1,j}(t^{\prime}+1) \frac{\partial h_{j}({t^{\prime}+1})}{\partial r_i (t^{\prime})}\\&+\gamma \sum_j {\E}^{\prime}_{2,j}(t^{\prime}) \frac{\partial [G_{j}^{\text {out }}(t^\prime)+\epsilon_j^2(t^\prime) \Delta_{j}^{\text {out }}(t^\prime)]}{\partial r_i (t^{\prime})} \\
& =-\gamma {\E}^{\prime}_{1,i}(t^\prime)+\gamma f^{\prime}\left(r_i (t^{\prime})\right) \sum_j {\E}^{\prime}_{1,j}(t^{\prime}+1)\mu_{j i}\\
&+\gamma \sum_j \E^{\prime}_{2,j}(t^{\prime}) \mu_{ji}^{\rm out}f^{\prime}(r_i(t^{\prime}))+\gamma \sum_j \E^{\prime}_{1,j}(t^\prime +1) \hat{\epsilon}^1_{ji}+\gamma \sum_j \E^{\prime}_{2,j}(t^{\prime})  \hat{\epsilon}^2_{ji},\\
\end{aligned}
\end{equation}
where $\gamma$ indicates the learning rate for the inference phase (we choose $\gamma = 0.1$ for all tasks), $\hat{\epsilon}^1_{ji} = \epsilon^1_j\left(t^{\prime}+1\right) \frac{\left(\varrho_{j i}-\left(\mu_{j i}\right)^2\right)f^{\prime}\left(r_i (t^{\prime})\right)f\left(r_i (t^{\prime})\right)}{\sqrt{\left(\Delta_{j}^{\mathrm{in}}(t^\prime +1)\right)^2+\left(\Delta_{j}^{\mathrm{rec}}(t^\prime)\right)^2}}$, and $\hat{\epsilon}^2_{ji} = \epsilon^2_j\left(t^{\prime}\right) \frac{\left(\varrho^{\rm out}_{j i}-\left(\mu^{\rm out}_{j i}\right)^2\right)f^{\prime}\left(r_i (t^{\prime})\right)f\left(r_i (t^{\prime})\right)}{\Delta_{j}^{\rm out}(t^{\prime})}$. It is evident that the last two terms in Eq.~\eqref{inference} are related to the fluctuations caused by the network statistics. The interplay between the network statistics and the prediction errors governs the belief dynamics, which was not considered in previous studies.
We emphasize this intrinsic property of neural dynamics is due to ongoing fluctuations of synaptic weights in the presence of circuit noise~\cite{Noise-2008}.
Equation~\eqref{inference} thus addresses how the neural belief is shaped under the fluctuating circuit environment.

The goal of this inference process is to find the best configuration of belief for synaptic weight modifications (aforementioned neuroplasticity).
When the decrease of the energy $\mathcal{F}$ becomes stable, e.g., $|\mathcal{F}^t - \mathcal{F}^{t-1}|<0.1$, or when a maximal number of iterations ($n$ in our algorithm~\ref{Alg}) is approached, the learning phase starts, i.e., the hyperparameters $[\mathrm{m}^{\ell}, \boldsymbol{\pi}^{\ell}, \boldsymbol{\Xi}^{\ell}]$ are updated based on the local error signal $\mathbf{\E}^{\prime}_{j}(t)$ with the following increments,
\begin{equation}\label{learning}
	\begin{aligned}
		\Delta {m}^\ell_{ij}& =-\eta\frac{\partial\mathcal{F}}{\partial m^\ell_{ij}}=  -\eta \sum_{t}\E^{\prime}_{\ell^{\prime},i}(t) \left[-\frac{1}{N_{\ell}} \xi_j^\ell-\epsilon_i^{\ell^{\prime}}(t) \frac{m_{i j}^{\ell} \pi_{i j}^{\ell}\left(\xi_j^\ell\right)^2}{(N^\ell)^2(1-\pi_{i j}^{\ell})\sqrt{\Delta_i^{\ell^{\prime}}}}\right],\\
		\Delta {\pi}^\ell_{ij}& = -\eta\frac{\partial\mathcal{F}}{\partial \pi^\ell_{ij}}= -\eta \sum_t \E^{\prime}_{\ell^{\prime},i}(t)\left[-\epsilon_i^{\ell^{\prime}}(t)\frac{\left(m_{i j}^{\ell}\right)^2 \left(\xi_j^\ell\right)^2}{2(N_\ell)^2\left(1-\pi_{i j}^{\ell}\right)^2 \sqrt{\Delta_i^{\ell^{\prime}}}}\right],\\
		\Delta {\Xi}^\ell_{ij}& =-\eta\frac{\partial\mathcal{F}}{\partial \Xi^\ell_{ij}} =-\eta \sum_t \E^{\prime}_{\ell^{\prime},i}(t) \left[-\epsilon_i^{\ell^{\prime}}(t)\frac{\left(\xi_j^\ell \right)^2}{2N_\ell \sqrt{\Delta_i^{\ell^{\prime}}}}\right],
	\end{aligned}
\end{equation}
where $\eta$ denotes the learning rate for the learning phase, $\Delta_i^1 = \left(\Delta_i^{\mathrm{in}}(t)\right)^2+\left(\Delta_{i}^{\mathrm{rec}}(t-1)\right)^2$ and $\Delta_i^2 = (\Delta^{\text{out}}_i(t))^2$. To derive Eq.~\eqref{learning}, the chain rule and mean-field dynamics [Eq.~\eqref{MFD}] are used. The meaning of superscripts depends on the network structure where the computation is carried out.  If $\ell = \text{in}$, $\ell^{\prime} = 1$, $\xi^{\ell}_j = x_j(t)$, $N_{\ell} = N_{\rm in}$; if $\ell$ indicates the recurrent reservoir, $\ell^{\prime}= 1$, $\xi^{\ell}_j = f(r_j(t-1))$,  $N_{\ell} = N$; if $\ell = \text{out}$, $\ell^{\prime} =2$, $\xi^{\ell}_j = f(r_j(t))$,  $N_{\ell} = N$. For an easy comprehension, we summarize all mathematical items and associated explanations in appendix~\ref{app-tab}.
The dynamics of $\pi$ and $\Xi$ is purely driven by the synaptic fluctuation, while the $m$ dynamics is contributed by the activity (belief or sensory observation) and the synaptic fluctuation. The $m$ yields impacts on $\pi$ and $\Xi$ as well. Note that the vanilla predictive coding does not take into account synaptic fluctuations (see also appendix~\ref{app-pccode}), which is indeed ubiquitous in neural circuits~\cite{Taro-2021}. One typical source is that the synaptic noise results from noisy biochemical processes underlying synaptic transmission, while the other source is the fluctuation of spine sizes in the neocortex, and the existence of silent synapses~\cite{Barb-2007,Huang-2018}. 

In practice, implementation of the meta learning rule in Eq.~\eqref{learning} follows immediately the inference phase, where the update of belief $\mathbf{r}$ has made $\mathcal{F}$ converge. To improve the prediction performance, the inference and learning phases are repeated a number of times. Prediction phase is carried out after a round of inference-learning loop, to test the model's generalization performance. Three phases can be concisely represented by the pseudocode in Alg.~\ref{Alg}. Codes to reproduce the numerical results provided in the next section are available in our GitHub~\cite{code}.
\begin{algorithm}[H]
	\caption{Meta predictive learning algorithm} \label{Alg}
	\begin{algorithmic}[1]
		\State \textit{\# Inference}
		\State Given: input $\mathbf{x}$, label $\hat{\mathbf{y}}$, randomly initialized belief $\mathbf{r}, \mathbf{r}_y = \hat{\mathbf{y}}$, standard Gaussian variables $\boldsymbol{\epsilon}^1$ and $\boldsymbol{\epsilon}^2$
		\For {${\rm iter}= 1,\ldots, n$}
		\For {$t = 1,\ldots, T$}
		\State {${h_{i}(t+1) =G_{i}^{\mathrm{rec}}(t)+G_i^{\mathrm{in}}(t+1)+\epsilon_i^{\mathrm{1}}(t+1) \sqrt{\left(\Delta_i^{\mathrm{in}}(t+1)\right)^2+\left(\Delta_{i}^{\mathrm{rec}}(t)\right)^2}}$};
		\State $y_k(t) = \phi\left(G_{k}^{\text {out }}(t)+\epsilon_k^2(t) \Delta_{k}^{\text {out }}(t)\right)$;
		\State $\mathbf{r}(t) =\mathbf{r}(t)+\Delta \mathbf{r}(t)$.
		\EndFor
		\EndFor
		\State \textit{\# Learning}
		\For {$\ell = \text{in}, \text{out},\text{recurent}$}
		\For {$t = 1,\ldots, T$}
		\State $\mathbf{m}^{\ell} =\mathbf{m}^{\ell}+ \Delta \mathbf{m}^\ell$;
		\State $\boldsymbol{\pi}^{\ell} =\boldsymbol{\pi}^{\ell}+ \Delta\boldsymbol{\pi}^\ell$;
		\State $\boldsymbol{\Xi}^{\ell} =\boldsymbol{\Xi}^{\ell}+ \Delta\boldsymbol{\Xi}^\ell$.
		\EndFor
		\EndFor
		\State Output: $\mathbf{r}$.
		\State \textit{\# Prediction}
		\State Given: test data $\mathbf{x}$, converged belief $\mathbf{r}$, another set of standard Gaussian variables $\boldsymbol{\epsilon}^1$ and $\boldsymbol{\epsilon}^2$
		\For {$t = 1,\ldots, T$}
		\State {${h_{i}(t+1) =G_{i}^{\mathrm{rec}}(t)+G_i^{\mathrm{in}}(t+1)+\epsilon_i^{\mathrm{1}}(t+1) \sqrt{\left(\Delta_i^{\mathrm{in}}(t+1)\right)^2+\left(\Delta_{i}^{\mathrm{rec}}(t)\right)^2}}$};
		\State $y_k(t) = \phi\left(G_{k}^{\text {out }}(t)+\epsilon_k^2(t) \Delta_{k}^{\text {out }}(t)\right)$;
		\EndFor
		\State Output: ${\mathbf{y}}$.
	\end{algorithmic} 
\end{algorithm}

\section{Results and discussion}
In this section, we first apply the MPL in the digit classification task, where an MNIST digit of $784$ pixels is divided into a sequence of pixels, and subgroups of $28$ pixels are input to the network at each single time step. As a proof of concept, the first example is to show our framework can be applied to any computational tasks of temporal structures. Then, we extend the application to two language processing tasks; one is at the toy level and the other is the real corpus.
\subsection{MNIST digit classification}
\begin{figure}
	\centering
	\includegraphics[scale = 0.8]{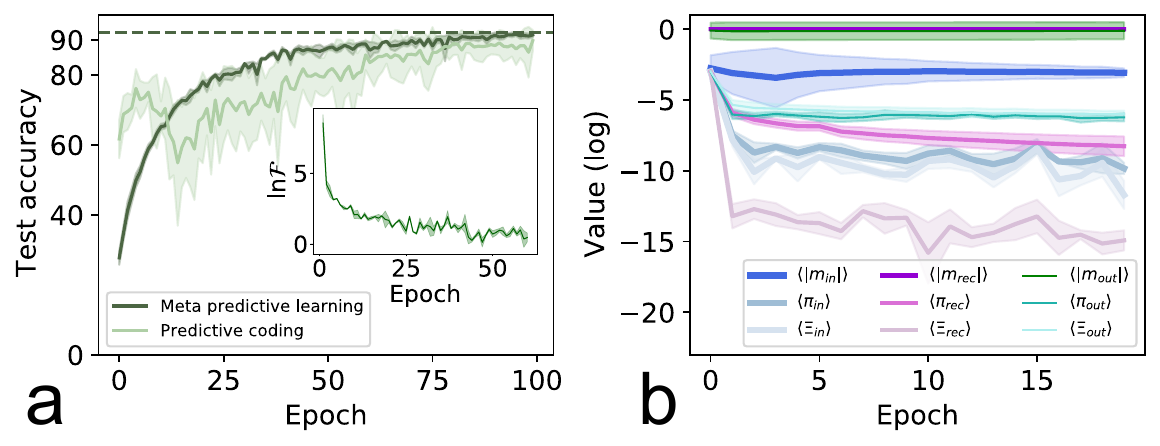}
	\caption{The performance of meta predictive learning on the 28 by 28 MNIST classification task. (a) Test accuracy as a function of epoch. The network with $N=100$ recurrent neurons, $N_{\rm in}=28$ input units and $N_{\rm out} = 10$ output nodes is trained on the full MNIST dataset with $60$k training images (handwritten digits), and validated on another unseen $10$k test handwritten digits. Predictive coding indicates the learning direct in the weight space rather than the distribution space. If the epoch is less than 40, the number of inference steps is set to $n=100$, and $n=200$ otherwise. The inset shows how $\ln \mathcal{F}$ changes with training in the first 60 training epochs (this log-energy becomes stable in the late training stage, and is thus not shown). Five independent runs are considered for the fluctuation of the result. (b) The logarithmic average value of $[\bm{\Xi}^\ell, \bm{\pi}^\ell,\mathbf{m}^\ell]$ versus epoch in all layers, the log means logarithm with the base $e$. Only the first twenty epochs are considered (the result remains stable in the later training stage), and the fluctuation is computed from five independent runs.} 
	\label{fig1}
\end{figure}

The recurrent neural network is trained to classify an input image after 28 time steps, seeing 28 pixels at each time step. This task requires long-term memory,  because the recurrent neural network makes the final decision only after seeing all the pixels, and the information in the previous time steps (up to 28 steps before) must be stored and processed in the last step. To carry out this task, we use a vanilla RNN with $N=100$ recurrent neurons, $N_{\rm in}=28$ input units and $N_{\rm out} = 10$ output nodes indicating the output class in the one-hot form. The entire dataset is divided into several batches, and we use stochastic gradient descent (SGD) in the learning phase to update hyperparameters $[\mathbf{m}^\ell,\boldsymbol{\pi}^\ell, \boldsymbol{\Xi}^\ell]$, and adam optimizer is applied~\cite{adam}. Despite working on the network ensemble level and the fact that weight uncertainty must be taken into account during inference, learning and prediction phases, our model can achieve better and more stable performances than the predictive coding without any distribution training [Fig.~\ref{fig1} (a)]. As expected, the overall energy $\mathcal{F}$ consistently decreases during training and reaches the point near zero in the late stage of training.

The macroscopic behavior of the network is corroborated by the statistical pattern of model hyperparameters underlying synaptic weights.
 The weight uncertainty characterized by hyperparameters $[\bm{\Xi}^\ell, \bm{\pi}^\ell]$ decreases over training, showing that the weight is becoming more deterministic, and we use the average value, e.g., $\langle \Xi_{\rm in} \rangle = \frac{1}{N\times N_{\rm in}}\sum_{ij}\Xi_{ij}^{\rm in}$ to compute the average uncertainty level (for the mean $\mathbf{m}$, we take its absolute value before the average is carried out). Interestingly, the uncertainty level is highest in the output layer, which is in striking contrast to the results obtained by a generalized backpropagation through time (rather than local learning guided by prediction error considered in the current work) at the ensemble level~\cite{Zou-2023} where the uncertainty is highest in the recurrent layer.  From the predictive coding perspective, the readout weight has more flexibility to extract the information in the reservoir, which may be due to the local nature of learning that is driven by minimizing the prediction error. This remarks that a  more biological plausible training may lead to different interpretations of the same computational tasks as implemented in neural circuits. Therefore, to reveal biological mechanisms, a biological plausible training is a necessary ingredient.

\subsection{Toy language model}
Real language corpus is commonly complicated, and is not simple for theoretical studies. To build a metaphor of the complicated natural language, we set up a generative process where a text (a stream of tokens) is generated through a fixed rule (similar to grammar). Following this setting, 
the artificial corpus consists of $M$ texts of length $T$ each, and each text is composed of letters from $a, b, c, \ldots, z$. A periodic boundary is applied. For example, a single sample $x = \{a,c,g,i, ...\}$ is generated according to the grammatical rule that starting from letter $'a'$, only the letter $'c'$ or $'e'$ which is located two letters or four letters next to $'a'$ (with equal probabilities) can follow $'a'$, and the case of two consecutive $'c'$ is not allowed. This rule for generating toy language is just a simple model of real corpus, but non-trivial enough for a neural network to learn the embedded rule.  The generated examples (letter sequences) are shown to the neural network, which is required to discover the rule by our meta predictive learning working on next-letter prediction. After training, the network is tested by generating a sequence of arbitrary length following the same rule. A hierarchical compositional structure can also be incorporated into the generation process, but we leave this more interesting case to future studies based on this toy setting.

\begin{figure}
	\centering
	\includegraphics[scale = 0.4]{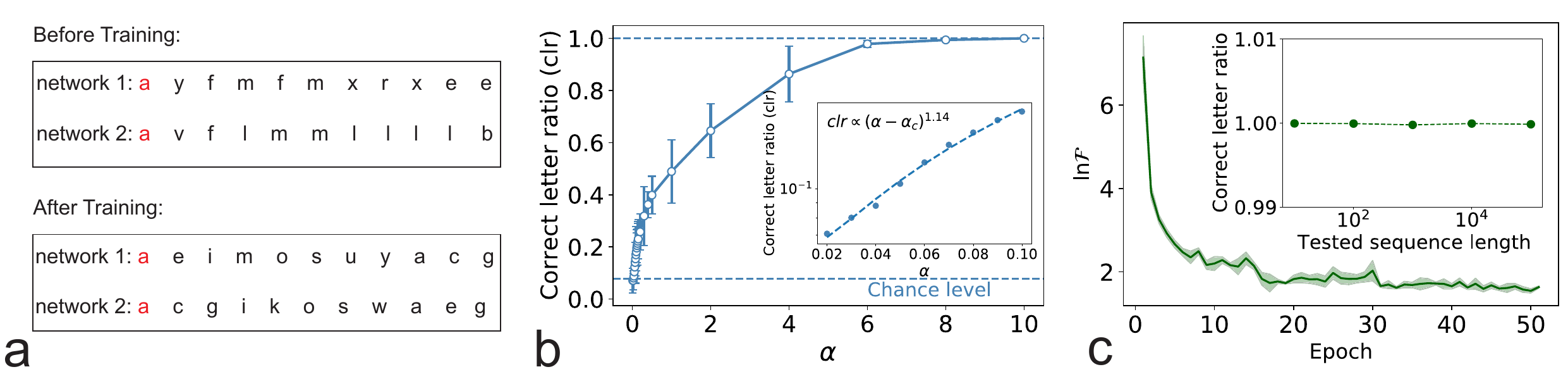}
	\caption{The properties of meta predictive learning on the simplified language prediction task. The grammatical rule is designed as follows: starting from a random letter ($'a'$ here), only the candidates located two letters or four letters after $'a'$ can follow the starting letter with equal probability, and each letter only repeats once in this next-word generation. All letters in the alphabet form a cyclic structure. $T=11$ is considered, and the full size of dataset is $26624$. RNN with $N=100, N_{\rm in}=26, N_{\rm out} = 26$ is trained, and two instances of networks are randomly sampled from the (trained or untrained) network ensemble. (a) Starting from the letter a, the network generates the next letter which serves as the input at the next time step, until a sequence with desired length is generated. (b) The correct letter ratio as a function of data load $\alpha = \frac{M}{N}$, and five independent runs are considered. $M$ examples of sequences are used for training. A chance level of $\frac{1}{13}$ is marked. The inset shows the correct letter ratio in the range of $\alpha\in [0.02,0.1]$. (c) The log-energy $\ln \mathcal{F}$ changes with training epochs and decreases to near zero. The inset shows how the correct letter ratio changes with the length of generated sequence after a full dataset is used for training. The error bar is computed with five independent networks.} 
	\label{fig2}
\end{figure}

 A RNN with $N=100, N_{\rm in}=26, N_{\rm out} = 26$ is trained on the full dataset following the above rule, with a total of $26624$ (calculated as $26\times2^{T-1}$) sequences of length $T=11$ (other values of $T$ can also be similarly studied), and SGD with adam optimizer is applied~\cite{adam}. To detect a possible phase transition with increasing data size, we can use an increasing portion of the entire dataset (i.e., $M<26624$). Each letter can be encoded into one-hot form before being input to the network, while the readout implements a decoding of the neural activity into one-hot form as well. Because of simplicity in our letter space, we do not need word embedding as commonly used in language processing~\cite{WB-2013}. In Fig.~\ref{fig2} (a), we can easily generate a sequence of arbitrary length, by supplying a network with the letter generated in the previous time step, and the trained network (in the ensemble sense) successfully generates sequences following the ground truth rule. Interestingly, the well-trained network also generates sequences with length $T>11$ following the same rule, suggesting the possibility that the network output could be creative to generate new grammatically correct sequences.

\begin{figure}
	\centering
	\includegraphics[scale = 0.65]{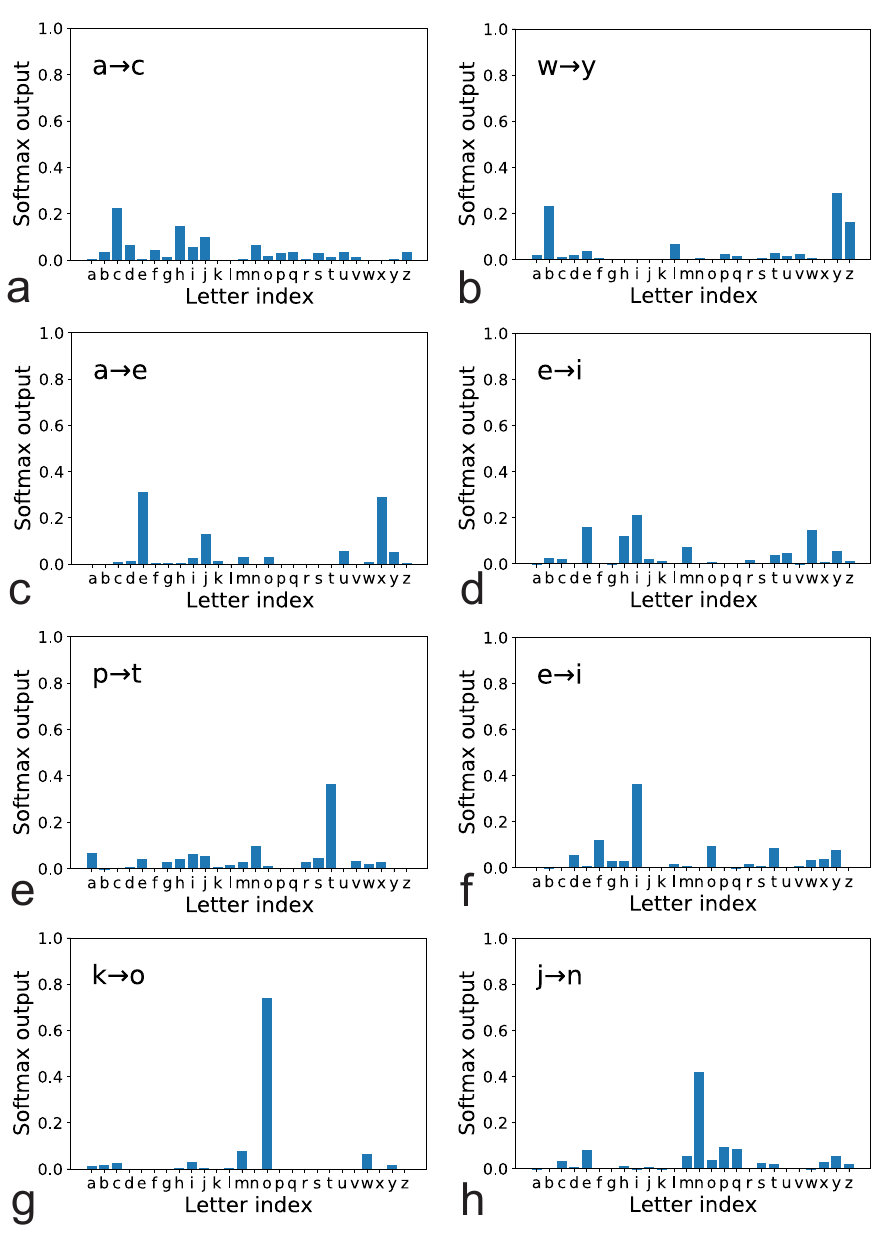}
	\caption{ Softmax values of the output units for different data load $\alpha$. Panels (a,b), (c,d), (e,f) and (g,h) show two typical patterns for each data load $\alpha = 0$, $\alpha = 0.01$, $\alpha = 0.03$, and $\alpha = 0.05$, respectively. Only predictions following the designed language rule are displayed, and the text shown in the panel $''a\to c''$ means inputting the letter $'a'$ and the network predicts the immediate following letter $'c'$ (corresponding to the largest softmax output). The training conditions are the same as in Fig.~\ref{fig2}.} 
	\label{vis}
\end{figure}

To study the emergence behavior of this simplified language model, we define the correct letter ratio to characterize the language generating ability of our model. After training, the network instance (sampled from the ensemble) is required to generate 26 sequences of length $T=11$ whose first letters are one of all 26 letters of the alphabet, and the correct letter ratio is defined as the average ratio of correctly predicted letters. For example, the sequence $['a', 'c', 'e', 'g', 'k', 'm', 'o', 's', 'w', 'a', 'z']$ has $9$ correctly predictions, with ratio $0.9$ (in total the network has to predict 10 letters) for this single sequence. Therefore, the correct letter ratio indicates the language generating ability of the network ensemble, with a maximal value of $1$ ($100\%$). In Fig.~\ref{fig2} (b), we can easily see that the correct letter ratio first remains at a very low level (close to chance level) if the data load $\alpha = \frac{M}{N}$ is small, i.e.,  the generated sequences are random when $\alpha <0.02$. Beyond this threshold, the performance continuously improves, exhibiting the phenomenon of a second-order phase transition, which coincides qualitatively with empirical results of emergence discovered in large language models~\cite{Kaplan-2020,Wei-2022}. The scaling exponent of the correct letter ratio (order parameter in statistical mechanics~\cite{Huang-2022}) around the transition point is about $1.14$. 
A rigorous derivation of this critical exponent is left for future analytic works. Training RNN with different network sizes yields qualitatively same behavior, but a larger network size makes the transition sharper.
After the transition, the network assigns the correctly predicted letter with a larger probability than other letter candidates, while the possibilities for other letters are significantly suppressed (see Fig.~\ref{vis}). Another important characteristic is that, the learning with increasing data occurs first rapidly, followed by a slow period, and finally the performance is saturated to the perfect generalization of the language rule. This may be interpreted as a hierarchical decoding of the information embedded in the noisy (stochasticity in the generation process) sequences.  We further remark that, after a full dataset of sequences with fixed length (e.g., $T=11$) is trained, the network is able to generate the grammatically correct letter sequences of \textit{arbitrary} length [see the inset of Fig.~\ref{fig2} (c)]. The energy of the language model is also decreasing with training until getting stationary, which emphasizes the important role of energy-based model to understand recurrent language processing. A further extension of meta predictive learning to transformer structure is possible, as the Gaussian assumption used in the standard predictive coding can be gone beyond in a recent work~\cite{Beyond-2022}.
\begin{figure}
	\centering
	\includegraphics[scale = 0.4]{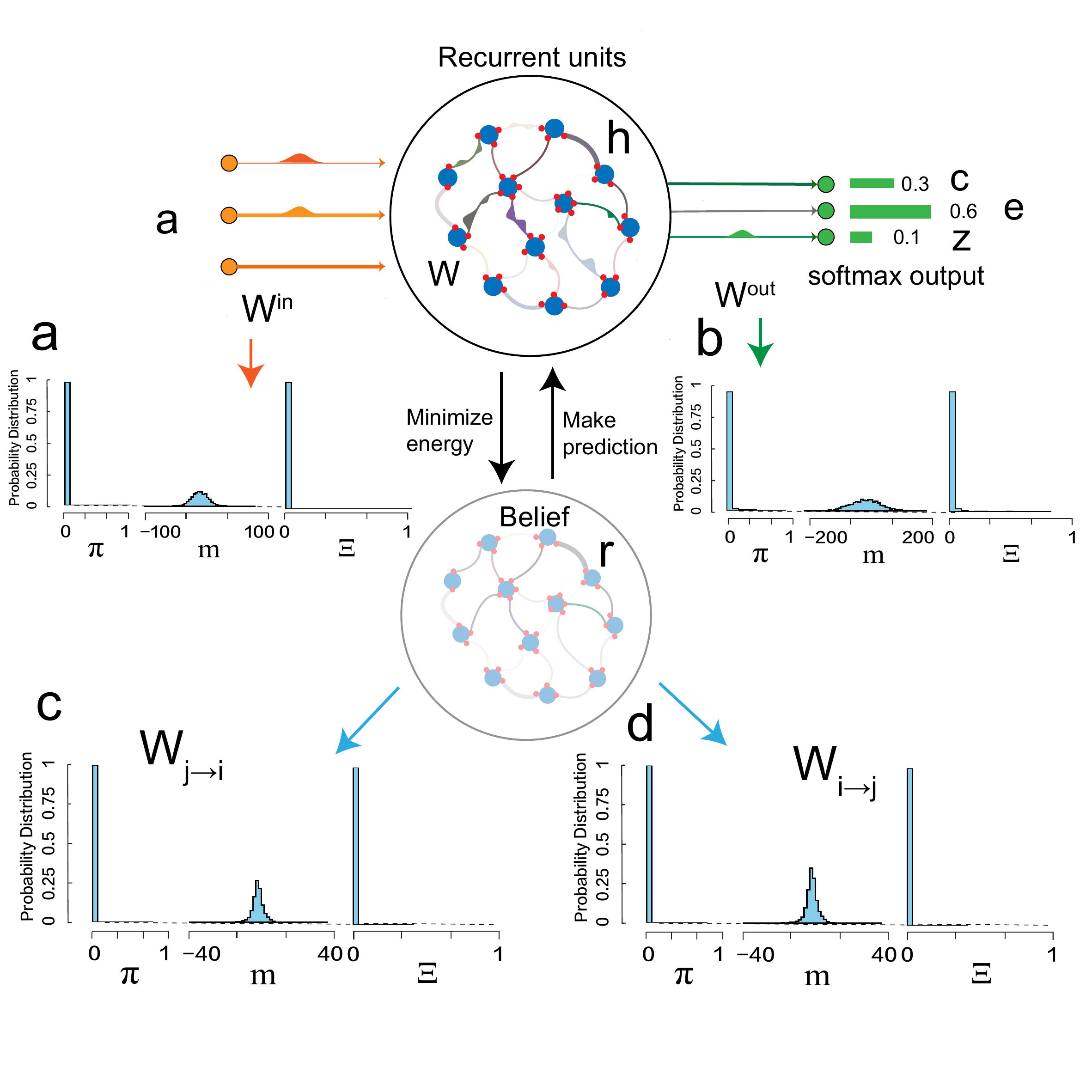}
	\caption{Illustration of hyperparameters $[\pi, m, \Xi]$ in meta predictive learning on the simplified language task. 
	The training conditions are the same as in Fig.~\ref{fig2}. In (c-d), we show statistical properties of bidirectional connections, and $i<j$ is considered.} 
	\label{fig3}
\end{figure} 

To study the properties of this simplified language model, we plot the distribution of hyperparameters $[\pi, m, \Xi]$  for the input layer, output layer, and recurrent layer, respectively. The distribution of $[\pi, \Xi]$ has the L-shape in all layers, while the output layer allows for more variability in both sparsity and variance of the Gaussian slab, which is characterized by a slightly broader distribution of $[\pi, \Xi]$. Extremes $\pi = 0, \pi =1$ and $\Xi = 0$ have particular physics significance. $\pi = 0$ indicates the connection has no sparsity, and thus carries important information for the task. The spike mass at $\pi = 1$ implies that the connection is always zero, and thus is not important for the task, but none of the connections of our model belong to this case. $\Xi = 0$ shows the corresponding connection is deterministic, because the corresponding Gaussian distribution reduces to a Dirac delta peak. This result is also observed in the 28 by 28 MNIST classification task. The distribution of hyperparameter $m$ is broadest in the output layer, ranging from $-200$ to $200$, showing the higher-level variability in the connection weight of the output layer. This phenomenon may have close relationship with the fact that the embedded rule can only be retrieved by using a highly heterogeneous weighting of each neuron's activity in the reservoir, which is particularly interesting from the perspective of neural decoding of language information and probabilistic computation in a biological plausible setting~\cite{Pouget-2013,Wang-2023,Tenen-2023}, since our embedded rule is actually a probabilistic generative rule mixed with a predefined grammatical structure.  

\begin{figure}
    \centering
    \includegraphics[scale=0.7]{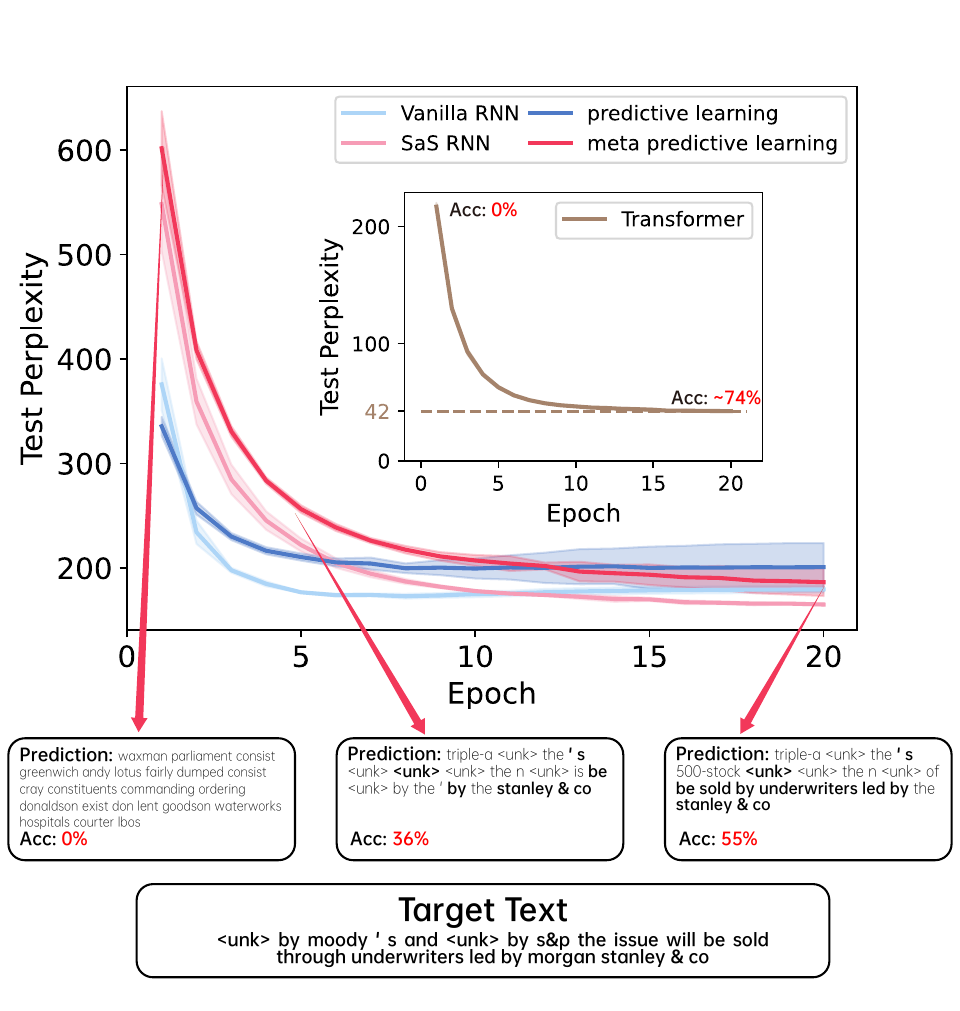}
    \caption{Training performance of networks with different architectures in Penn Treebank dataset. In the upper part of the figure, we choose the vanilla RNN~\cite{Huang-2022}, SaS RNN (ensemble learning)~\cite{Zou-2023}, RNN with standard predictive coding~\cite{Exact-2021} and RNN with meta predictive learning to show how test perplexity decreases with the training epoch. The first two algorithms belong to the backpropagation through time category~\cite{Huang-2022}. In the inset, we provide the performance of transformer model (see details in appendix~\ref{app-transformer}) with single encoder block for comparison. 
    We also mark the mean test accuracy of the transformer model at the beginning of training and at the end of the training.
    In the bottom part of the figure, we select untrained, trained-for-five-epoch, and full-trained RNN with meta predictive learning to show the performances at different training stages in generating one of the sentences in the test dataset. The correctly predicted tokens from the test sentence are highlighted, while the wrongly predicted tokens are gray colored. The indicated accuracy is the ratio of the number of correctly predicted tokens from the test sentence to the total number of tokens in the sentence. The mean accuracy evaluated from $100$ sentences is about $0\%$, $21.3\%\pm10.5\%$, $23.5\%\pm11.3\%$ at the three shown stages, respectively. Note that all the models share the same training hyperparameters like batch size, learning rate, and training optimizers (see appendix~\ref{app-transformer} for details).}
    \label{fig: learningCurvePTB}
\end{figure}

\begin{figure}
    \centering
    \includegraphics[scale=0.65]{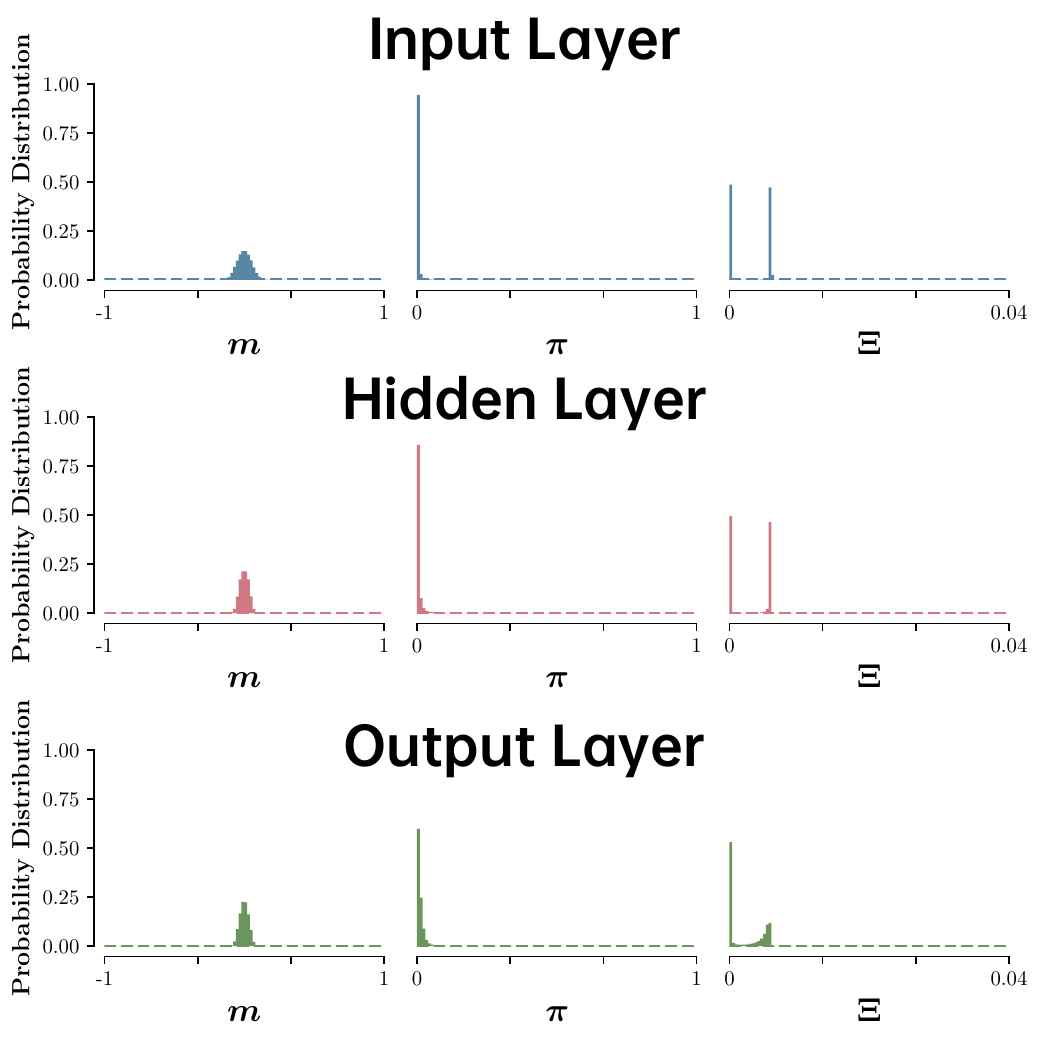}
    \caption{Probability distribution of hyperparameters $m$, $\pi$, $\Xi$ in the RNN networks trained with meta predictive learning. Distributions of hyperparameters $m$, $\pi$, $\Xi$ in the input layer are shown in the top of the figure (blue histogram), and those in the hidden layer and output layer are shown in the middle (salmon histogram) and in the bottom (green histogram) of the figure, respectively.}
    \label{fig: densityDistributionPTB}
\end{figure}

\subsection{Experiments on natural language}
In this section, we apply our MPL algorithm to a more complex language corpus, i.e., Penn Treebank (PTB) corpus~\cite{data-1993}, which contains nearly 50000 sentences collected from the Wall Street Journal. The PTB is one of the most known and used corpus for word-level language modeling. 

Due to the large space of the corresponding vocabulary, the corpus need to be pre-processed using word embedding techniques before sending the sentences into the network~\cite{WB-2013}. Here we describe the main steps.
The first step is to use a tokenizer tool to split the sentences into tokens and replace the useless words or characters with a special token named $<$unk$>$, indicating an unknown token. In addition, the tokens that appeared less than five times in the whole corpus will be replaced with $<$unk$>$ to help the network concentrate on the major tokens of high frequency.
Next, we collect all tokens to generate a vocabulary to store all the different tokens. 
However, directly inputting the tokens (treated as one-hot vectors) into the network is inconvenient when the size of vocabulary is large. 
Hence, we set up a look-up table called embedding layer, to transform every token into vectors in a low-dimensional feature space via neural networks. The training goal is to learn word vector representations that are able to predict the nearby words. Rows of the trained encoding matrix gives the distributed representation of words.
The embedding layer is trained by traditional back-propagation algorithm~\cite{WB-2013}, while both the recurrent reservoir and the readout layer are trained by our MPL as described above (or other alternatives if comparison among algorithms is made).

The overall energy for the predictive learning is given below,
\begin{equation}
    \mathcal{F} = \half \sum_t ||\E'_{\text{rec}}(t)||^2 + \mathcal{L},
\end{equation}
where $\E_{\rm rec}'$ denotes the prediction error for the recurrent reservoir, $\mathcal{L} (\mathbf{y}, \mathbf{r}_y) = -\sum_t \sum_i (\mathbf{r}_y)_i(t) \ln y_i(t)$ is related to the readout error.
In order to measure the accuracy of the language model, we use perplexity metric which measures how well the model predicts the next token, i.e., the uncertainty about the prediction, precisely given by~\cite{Bengio-2003}
\begin{equation}
    \text{ppl} = \left[\prod_{i = 1}^{T} p(w_i | w_{i - 1} ,\cdots ,w_0) \right]^{-\frac{1}{T}}.
\end{equation}
It is intuitive that minimizing the perplexity is equivalent to maximizing the probability of a corpus which is composed of $T$ words indicated by $\{w_0, w_1,\ldots,w_T\}$.
Because the output of our model $y_i(t)$ is actually the prediction probability in the non-linear softmax form, we can recast the perplexity as $\text{ppl}=e^{\mathcal{L}}$, where $\mathcal{L}$ represents the cross-entropy objective used to train the network.

We apply our MPL to model this real corpus with comparison among other competitive algorithms (see Fig.~\ref{fig: learningCurvePTB}). The test perplexity obtained by backpropagation through time with/without meta learning reaches a similar level to those obtained by standard predictive coding and our MPL method, after tens of epochs. A salient feature is that, those trainings without the SaS premise, get easily overfitted at later stages of training. For comparison, the transformer network with self-attention blocks (see appendix~\ref{app-transformer} for details), achieves the lowest perplexity among all considered methods,
which demonstrates that the current biological recurrent computation has still a large gap to the artificial transformer computation, where the input 
tokens are not shown to the network in sequence, but in a form of single block, such that the self-attention is able to integrate information from different parts of the input. This also implies that, new elements, e.g., attention mechanisms, possibly in to-be-revealed biological forms, might be added to our current framework, to minimize the gap on one hand, and on the other hand to develop a biological computational model of intelligent systems that can handle natural language, particularly without relying on long working memory and an astonishingly large corpus.

To study the network behavior, we plot the distribution of hyperparameters $m$, $\pi$, $\Xi$ when the RNN network is trained with the MPL method, as shown in the Fig.~\ref{fig: densityDistributionPTB}. We find that the mean weight $m$ for all layers is symmetrically distributed around zero, with a relatively narrow distribution. The distribution of $\pi$ for all layers is of an L-shape and peaks at $\pi=0$, indicating a dense network is favored and formed after learning. The distribution of $\Xi$ is of the U-shape and has two peaks. One peak is at $\Xi=0$, indicating that these weights are deterministic and could only take a single value of $m$, and the other peak is at $\Xi \simeq 0.01$, indicating that the corresponding connection can carry a range of candidate values. Currently, it remains unknown how to relate these microscopic details of the network structure to the decoding of the semantic information in the corpus. It is thus important in future works to design analytically tractable model of language processing bridging neurophysiological plausibility and superperformance observed in the state-of-the-art architectures, which would help to uncover key neuron, synapse, and circuit motif types in the human brain.

\section{Conclusion}
Predictive coding is a prediction-error driven learning with local updates, performing a joint process of both inference and learning, thereby being a potential candidate for how the brain builds an internal generative model of the complex evolving outside world~\cite{Huang-2023}.
We take the predictive coding within the language processing context, which is currently attracting an intense research interests due to ChatGPT~\cite{Sparks-2023}. We address a meta predictive learning mechanism in recurrent neural networks encoding and predicting tokens in text sequences, in the presence of uncertainty. A continuous phase transition is revealed in our model, and perfect generation can be achieved after a sufficient number of training sequences are provided. Therefore, our toy model provides a good starting point to dissect the mechanism of learning in language processing~\cite{Huang-2023}.

Our MPL framework is relatively not prone to overfitting in training real corpus. In addition, there emerge intriguing statistical patterns of hyperparameters in our networks. However, it remains unclear how these statistical properties explain the performance (e.g., accuracy in next-token predictions) of the recurrent computation., which highly resembles what occurs in human brains. In contrast, the self-attention leveraged in transformer networks is not biological (e.g., not recurrent and non-local learning). Nevertheless, the transformer structure leads to emergence of intelligence to some extent, and in particular the phenomenon of in-context learning, where the trained network can perform novel tasks by a prompt of example demonstrations without any further learning. The ability of in-context learning emerges by only scaling models and computation costs~\cite{Wei-2022}. The deviation from known brain computation for language processing triggers a hot debate on what the nature of intelligence is~\cite{Sej-2023}, and whether the intelligence can be achieved by next-token prediction~\cite{Tenen-2023}. More precisely, how a compressed representation of hierarchical compositional structure in linguistic data can be achieved by biological learning (or resulting in multi-task performances beyond transformer) remains largely mysterious. Our current study shows that meta predictive learning for language processing may be a fruitful route towards this goal.

A recent work demonstrated that the weight uncertainty with the form of SaS structure can be also incorporated into the transformer~\cite{Transas-2023}. In addition, gated recurrent neural networks with multiplicative mechanisms were recently shown to be able to learn to implement linear self-attention~\cite{Gate-2023}. Furthermore, the relationship between linear transformers allowing for faster autoregressive learning and RNNs was established in a recent work~\cite{ICML-2020}.  Taken together, our current work would be a starting point to establish the bridge between the biological learning (towards the science of specialized brain circuits) and transformer learning within the seminal predictive coding hypothesis, which can be put in the theoretically solid variational free energy minimization conceptual framework.

\begin{acknowledgments}
This research was supported by the National Natural Science Foundation of China for
Grant Number 12122515 (H.H.), and Guangdong Provincial Key Laboratory of Magnetoelectric Physics and Devices (No. 2022B1212010008), 
and Guangdong Basic and Applied Basic Research Foundation (Grant No. 2023B1515040023).  
\end{acknowledgments}

\appendix
\section{Interpretation of predictive coding as variational free energy minimization}
\label{app-pc}
For a recurrent dynamics, we write the neural activity at each step as a latent variable $\mathbf{r}(t)$, and then it is reasonable to assume the joint probability
of a trajectory can be written into the following Markovian form,
\begin{equation}
P(\mathbf{r}(0),\ldots,\mathbf{r}(T))=P(\mathbf{r}(0))\prod_{t=1}^{T}P(\mathbf{r}(t)|\mathbf{r}(t-1)).
\end{equation}
We further assume a Gaussian form for the transition probability $P(\mathbf{r}(t)|\mathbf{r}(t-1))=\mathcal{N}(\mathbf{r}(t);\mathbf{h}(t),\sigma^2_t\mathbf{I})$, where
$\mathbf{h}(t)=\mathbf{w}f(\mathbf{r}(t-1))$, and $\sigma^2_t$ is a time-dependent variance, and for simplicity, we can set the variance to one without loss of generality, as the mere effect is leading to a rescaled cost function below. This Gaussian form can be obtained as an approximation, by using the Laplace approximation even if the transition probability is of other forms.
The goal is to optimize the negative log-likelihood of the joint distribution, defined by
\begin{equation}
\mathcal{F}=-\ln P(\mathbf{r}(0),\ldots,\mathbf{r}(T))=\frac{1}{2}\sum_t\frac{\lVert\mathbf{r}(t)-\mathbf{h}(t)\rVert^2}{\sigma^2_t}+\text{const},
\end{equation}
which corresponds exactly to the cost function of predictive coding if we treat $\sigma^2_t=1$ and neglect the constant term.

\section{Transformer model}
\label{app-transformer}
A transformer network consists of an embedding layer, encoder blocks, and decoder blocks~\cite{Att-2017}. In analogy to the RNN model, all tokens (one-hot vectors) are transformed into representations $\mathbf{X} \in \mathbb{R}^{d\times T}$ by an embedding layer, where $d$ denotes the dimension of embedding space and $T$ denotes the sequence length. As a clear difference from the RNN training, the input to the transformer is an entire $\mathbf{X}$ matrix, rather than one column each step for RNN.
 Note that we have not considered the position encoding scheme (e.g., adding a vector of sinusoids of different frequencies and phases to encode position of a word in a sentence) in our model. 

An encoder block includes two parts. The first part is the self-attention mechanism, aiming to evaluate the correlations among words in the input block $\mathbf{X}$. To this end, we introduce three trainable matrices, namely, query $Q$, key $K$, and value $V$. Then, a linear transformation of the input is applied.
\begin{equation}
\begin{aligned}
    Q =& W_Q\cdot\mathbf{X}, \\
    K =& W_K \cdot\mathbf{X},\\
    V =& W_V\cdot\mathbf{X},
\end{aligned}
\end{equation}
where $W_Q, W_K \in \mathbb{R}^{d_h \times d}$ and $W_V \in \mathbb{R}^{d \times d}$ are transformation matrices, and $d_h$ is the internal size of the attention operation. Therefore, we define $X_t$ as the $t$-th column of $\mathbf{X}$, and then we can define three vectors, namely $k_t=W_KX_t$, $v_t=W_vX_t$, and $q_t=W_QX_t$. 
Then, the $t$-th column of the self-attention matrix $\operatorname{SA}(\mathbf{X})$ is given by 
\begin{equation}
\begin{aligned}
\operatorname{attn}(t)&=\sum_{i=1}^T\alpha_i(t)v_i,\\
\alpha_i(t)&=\frac{e^{k_i^\top q_t/\sqrt{d_h}}}{\sum_{j=1}^Te^{k_j^\top q_t/\sqrt{d_h}}},
\end{aligned}
\end{equation}
where $\alpha_i(t)$ is a softmax operation containing information about the pairwise interactions between tokens. The normalization factor $\sqrt{d_h}$ is required to retain 
relevant quantities in the exponential function being of the order one.

The second part is two feed-forward layers with skip connection, i.e.,
\begin{equation}
\begin{aligned}
    \mathbf{z}_1 =& \operatorname{SA}(\mathbf{X}) + \mathbf{X} & \qty(\text{residual layer 1})\\
    \mathbf{z}_2 =& \operatorname{ReLU} \qty(W_1\cdot\mathbf{z}_1  + b_1) & \qty(\text{feed-forward layer 1}) \\
    \mathbf{z}_3 =&W_2\cdot  \mathbf{z}_2  + b_2 & \qty(\text{feed-forward layer 2}) \\
    \mathbf{z}^{\text{out}} =& \mathbf{z}_1 + \mathbf{z}_3 & \qty(\text{residual layer 2})
\end{aligned}
\end{equation}
where $W_1, W_2$ and $b_1, b_2$ are weights and biases of the two feed-forward layers. The output representations $\mathbf{z}^{\text{out}}$ can be considered to be the input of the next encoder block. Here, we use the single headed attention transformer, and do not use the layer normalization, which scales each element of a vector by the mean and variance of all elements in that vector.

Our used transformer model  has only one encoder block and one decoder layer. The decoder layer is  a linear layer (the readout layer), where the output representations $\mathbf{z}^{\text{out}}$ can be translated into the probability of the next token, which have the same function as the readout layer of the RNN model.
The dimension of representations $d = 300$ for all models in Figure~\ref{fig: learningCurvePTB}. For four RNN models, the number of neurons in the recurrent reservoir is $N = 512$. For the transformer model, it is convenient to set the hidden dimension $d_h = d = 300$.
The training parameters for all models are set to be the same. The batch size is $128$ and the learning rate is $0.001$. We have chosen the Adam algorithm as our training optimizer~\cite{adam}.

\section{The vanilla predictive learning algorithm}\label{app-pccode}
The vanilla predictive learning algorithm is a simplified version of our meta-predictive learning algorithm, without considering weight uncertainty.
Hence, setting $\boldsymbol{\pi} = 0$ and $\Xi = 0$ in Eq.~\eqref{inference} and Eq.~\eqref{learning} in the main text leads to the following update equations for belief and weights.

\begin{equation}\label{eq:vanillaPredictiveLearningUpdateR}
    \Delta r_i(t') = -\gamma \E_{1, i}^\prime (t') + \gamma f'(r_i(t'))\sum_{j} \E_{1, j}^\prime (t' + 1) w_{j i} + \gamma f'(r_i(t')) \sum_j \E_{2, j}^\prime(t') w_{j i}^{\rm out},
\end{equation}
and
\begin{equation}\label{eq:vanillaPredictiveLearningUpdateW}
    \Delta w_{i j}^\ell = \frac{\eta}{N_\ell} \sum_t \E_{\ell', i}^\prime (t) \xi_j^\ell,
\end{equation}
where the definition of $\ell$, $\ell'$, $\xi_j^\ell$ and $N_\ell$ bear the same meaning as in the main text (see Table~\ref{tab:item}).
We present the pseudocode of the vanilla predictive learning algorithm in Alg.~\ref{pcc}.

\begin{algorithm}[H]
\setcounter{algorithm}{1}
\caption{Vanilla predictive coding algorithm}\label{pcc}
    \begin{algorithmic}[1]
        \State \# Inference
        \State Given: input $\mathbf{x}$, label $\hat{\mathbf{y}}$, randomly initialized belief $\mathbf{r}$, $\mathbf{r}_y = \hat{\mathbf{y}}$
        \For{$i = 1, \ldots ,n$}
            \For{$t = 1, \ldots, T$}
                \State $h_i(t)=\sum_{j=1}^{N} w_{i j} f\left(r_j(t-1)\right)+\sum_{j=1}^{N_{\text {in }}} w_{i j}^{\text {in }} x_j(t)$;
                \State $y_i(t)=\phi\left(\sum_{j=1}^N w_{i j}^{\text {out }} f\left(r_j(t)\right)\right)$;
                \State $\mathbf{r}(t)=\mathbf{r}(t)+\Delta \mathbf{r}(t)$.
            \EndFor
        \EndFor

        \State \# Learning
        \For{$\ell = \text{in}, \text{out}$, \text{recurrent}}
            \For{$t = 1, \ldots ,T$}
                \State $\mathbf{w}^\ell = \mathbf{w}^\ell + \Delta \mathbf{w}^\ell$.
            \EndFor
        \EndFor
        \State Output: $\mathbf{r}$

        \State \# Prediction
        \State Given: test data $\mathbf{x}$, converged belief $\mathbf{r}$
        \For{$t = 1,\ldots,T$}
            \State $h_i(t)=\sum_{j=1}^{N} w_{i j} f\left(r_j(t-1)\right)+\sum_{j=1}^{N_{\text {in }}} w_{i j}^{\text {in }} x_j(t)$
            \State $y_i(t)=\phi\left(\sum_{j=1}^N w_{i j}^{\text {out }} f\left(r_j(t)\right)\right)$
        \EndFor

        \State Output: $\mathbf{y}$
    \end{algorithmic}
\end{algorithm}

\section{Mathematical items used in the main text and associated explanations}\label{app-tab}
We list mathematical items used in the main text and associated explanations to help readers go through the paper smoothly, as shown in the table~\ref{tab:item}.
\begin{table}[bt]
\caption{\label{tab:item}Mathematical items used in the main text and associated explanations.}
\begin{tabular}{l l}
\hline
Item & Explanation     \\
\hline
$r_i(t)$  & \textit{belief} of neuron $i$ at time step $t$   \\
 $h_i(t)$  & prediction of neuron $i$ at time step $t$   \\
 $y_i(t)$  & activity of output unit $i$ at time step $t$   \\
$\hat{\mathbf{y}}$  & label of input   \\
$f(\cdot)$  & non-linear activation function for recurrent dynamics   \\
$\phi(\cdot)$  & non-linear activation function for network output   \\
$\ell$ & in, out or recurrent (no superscript or subscript) \\
$N_{\ell}$ & the number of neurons for $\ell$\\
$w_{ij}^{\ell}$ &  directed coupling from neuron $j$ to neuron $i$\\
$\mu_{ij}^{\ell}$  & first moment of $w_{ij}^{\ell}$: $\mu^{\ell}_{i j}=\frac{m^{\ell}_{i j}}{N_\ell}$  \\
$\varrho^{\ell}_{ij}$ & second moment of $w_{ij}^{\ell}$: $\varrho^{\ell}_{i j}=\frac{\left(m^{\ell}_{i j}\right)^2}{N_\ell^2\left(1-\pi^{\ell}_{i j}\right)}+\frac{\Xi^{\ell}_{i j}}{N_\ell}$\\
$G_i^{\mathrm{in}}(t+1)$ & input mean current: $G_i^{\mathrm{in}}(t+1)  =\sum_j \mu_{i j}^{\mathrm{in}} x_j(t+1)$\\
$G_{i}^{\mathrm{rec}}(t+1)$ & recurrent mean current: $G_{i}^{\mathrm{rec}}(t+1)  =\sum_j \mu_{i j} f\left(r_{j}(t+1)\right) $\\
$G_{k}^{\mathrm{out}}(t+1)$ & output mean current: $G_{k}^{\mathrm{out}}(t+1)  =\sum_j \mu_{k j}^{\text {out }} f\left(r_{j}(t+1)\right)$ \\
$\Delta_i^{\mathrm{in}}(t+1)$ & input fluctuation: $\sqrt{\sum_j\left(\varrho_{i j}^{\text {in }}-\left(\mu_{i j}^{\mathrm{in}}\right)^2\right)\left(x_j(t+1)\right)^2 }$\\
 $\Delta_{i}^{\mathrm{rec}}(t+1)$ & recurrent fluctuation: $\sqrt{\sum_j\left(\varrho_{i j}-\left(\mu_{i j}\right)^2\right)\left(f\left(r_{j}(t+1)\right)\right)^2}$\\
 $\Delta_{k}^{\text {out }}(t+1)$ & output fluctuation: $\sqrt{\sum_j\left(\varrho_{k j}^{\text {out }}-\left(\mu_{k j}^{\text {out }}\right)^2\right)\left(f\left(r_{j}(t+1)\right)\right)^2 }$\\
$\Delta_i^1$ & $\Delta_i^1 = \left(\Delta_i^{\mathrm{in}}(t)\right)^2+\left(\Delta_{i}^{\mathrm{rec}}(t-1)\right)^2$\\
$\Delta_i^2$ & $\Delta_i^2 = (\Delta^{\text{out}}_i(t))^2$\\
$\boldsymbol{\epsilon}^1, \boldsymbol{\epsilon}^2$ & standard Gaussian variables\\
$\E_1(t)$ & $\E_1(t) =\frac{1}{2}\|\mathbf{{r}}(t) - \mathbf{{h}}(t)\|^2$\\
$\E_{2}(t)$ & $\E_{2}(t) = - \sum_i{\hat{y}_i}(t)\ln\left({{y}_i}(t)\right)$\\
$\mathcal{F}$ & energy cost: $\mathcal{F} = \sum_{j=1}^2\sum_{t=1}^T\E_{j}(t)$\\
$\mathbf{\E}^{\prime}_1(t)$ & prediction error vector 1: $\mathbf{\E}^{\prime}_1(t) = \mathbf{{r}}(t) - \mathbf{{h}}(t)$ \\
$\mathbf{\E}^{\prime}_2(t)$ & prediction error vector 2: $\mathbf{\E}^{\prime}_2(t) = \hat{\mathbf{y}}(t) - \mathbf{{y}}(t)$\\
$\hat{\epsilon}^1_{ji}$ &  $\hat{\epsilon}^1_{ji} = \epsilon^1_j\left(t^{\prime}+1\right) \frac{\left(\varrho_{j i}-\left(\mu_{j i}\right)^2\right)f^{\prime}\left(r_i (t^{\prime})\right)f\left(r_i (t^{\prime})\right)}{\sqrt{\Delta_j^1}}$\\
$\hat{\epsilon}^2_{ji} $ & $\hat{\epsilon}^2_{ji} = \epsilon^2_j\left(t^{\prime}\right) \frac{\left(\varrho^{\rm out}_{j i}-\left(\mu^{\rm out}_{j i}\right)^2\right)f^{\prime}\left(r_i (t^{\prime})\right)f\left(r_i (t^{\prime})\right)}{\sqrt{\Delta_j^2}}$\\
$\gamma$ & learning rate for the inference phase\\
$\eta$ & learning rate for the learning phase\\
${\xi}_j^{\ell}$ & $\xi^{\text{in}}_j = x_j(t)$, $\xi_j = f(r_j(t-1))$, $\xi^{\text{out}}_j = f(r_j(t))$\\
$M$ & the number of training sequence examples\\
$\alpha$ & data load $\alpha = \frac{M}{N}$\\
$T$ & sequence length\\
\hline
\end{tabular}
\end{table}

\end{document}